\documentclass[10pt,twocolumn,letterpaper]{article}

\usepackage{iccv}
\usepackage{times}
\usepackage{epsfig}
\usepackage{graphicx}
\usepackage{amsmath}
\usepackage{amssymb}
\usepackage{multirow}
\usepackage{makecell}
\usepackage{booktabs}
\usepackage{bbding}
\usepackage{url}
\usepackage{color}
\usepackage{caption}
\definecolor{orange}{rgb}{1, 0.5, 0}
\usepackage[pagebackref=true,breaklinks=true,letterpaper=true,colorlinks,bookmarks=false]{hyperref}

\iccvfinalcopy 

\def\iccvPaperID{****} 

\ificcvfinal\fi

\begin{document}

\title{Towards Universal Vision-language Omni-supervised Segmentation}

\author{Bowen Dong$^{1}$ \quad
Jiaxi Gu$^{2}$ \quad
Jianhua Han$^{2}$ \quad
Hang Xu$^{2}$ \quad
Wangmeng Zuo$^{1}$\textsuperscript{\Envelope} \\
$^1$Harbin Institute of Technology  \quad $^2$Huawei Noah’s Ark Lab \\
\texttt{\{cndongsky,imjiaxi,chromexbjxh\}@gmail.com} \\ \texttt{hanjianhua4@huawei.com, wmzuo@hit.edu.cn} \\
}
\ificcvfinal\thispagestyle{empty}\fi
\twocolumn[{%
\renewcommand\twocolumn[1][]{#1}%
\maketitle
\vspace{-2em}
\begin{center}
   \captionsetup{type=figure}
   \includegraphics[width=0.99\textwidth]{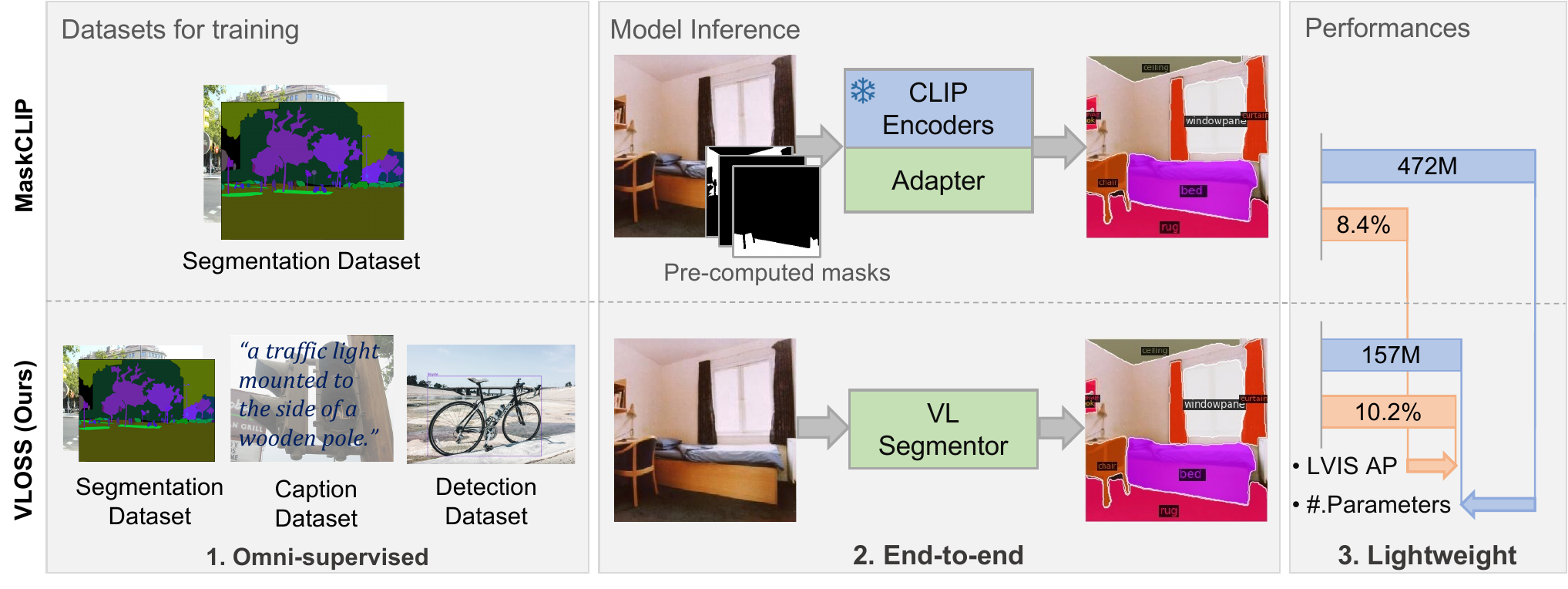}
   \vspace{-1em}
   \captionof{figure}{Comparison between two open-world universal segmentation methods: \textbf{(up)} MaskCLIP~\cite{ding2022open}, and  \textbf{(bottom)} our VLOSS. Compared to~\cite{ding2022open}, VLOSS has three merits: \textbf{1)} enabling omni-supervised training; \textbf{2)} can be optimized and inferenced in an end-to-end manner; and \textbf{3)} more lightweight yet better performance even without pretrained CLIP encoders.
   }\label{fig:overview}
\end{center}%
}]

\begin{abstract}
   %
   Existing open-world universal segmentation approaches usually leverage CLIP and pre-computed proposal masks to treat open-world segmentation tasks as proposal classification. 
   However, 1) these works cannot handle universal segmentation in an end-to-end manner, and 2) the limited scale of panoptic datasets restricts the open-world segmentation ability on things classes. 
   In this paper, we present Vision-Language Omni-Supervised Segmentation (VLOSS). 
   VLOSS starts from a Mask2Former universal segmentation framework with CLIP text encoder. To improve the open-world segmentation ability, we leverage omni-supervised data (\ie, panoptic segmentation data, object detection data, and image-text pairs data) into training, thus enriching the open-world segmentation ability and achieving better segmentation accuracy. To better improve the training efficiency and fully release the power of omni-supervised data, we propose several advanced techniques, \ie, FPN-style encoder, switchable training technique, and positive classification loss.  
   Benefiting from the end-to-end training manner with proposed techniques, VLOSS can be applied to various open-world segmentation tasks without further adaptation. 
   Experimental results on different open-world panoptic and instance segmentation benchmarks demonstrate the effectiveness of VLOSS. Notably, with fewer parameters, our VLOSS with Swin-Tiny backbone surpasses MaskCLIP by $\sim$2\% in terms of mask AP on LVIS v1 dataset.
\end{abstract}

\vspace{-2.5em}
\section{Introduction}\label{sec:intro}
\vspace{-0.5em}


%
Recently, vision-language-based (VL-based) pretrained models have achieved remarkable progress in various image recognition~\cite{zhou2022learning,zhou2022conditional,zhang2022tip} and scene understanding tasks~\cite{lin2022frozen,rao2022denseclip,zhao2022exploiting,liu2022open}. 
Notably, Vision-Language Pretraining typically represented by CLIP~\cite{radford2021learning,yao2022filip} introduces contrastive learning between massive image-text pairs to obtain well-aligned visual and textual representations simultaneously. 
Benefiting from the highly structured and semantically rich text data, as well as the massive amount of paired image-text data, the text encoder of CLIP learns abundant visual concepts and definitions, 
thus making CLIP pretrained model obtain the open-world recognition ability on image level. 

Nevertheless, most vision-language pretrained models~\cite{radford2021learning,yang2022unified,jia2021scaling} are learned by contrastive learning with low-resolution images. 
Without instance-level annotations and high-resolution training data, the vision-language models have to leverage off-the-shelf proposal generator to extract small regions with objects, then conduct object detection or segmentation~\cite{gu2022openvocabulary,wu2022end,xu2022semseg}, which restricts the instance-level recognition ability of vision-language pretrained models. 
To tackle this issue, a couple of recent works~\cite{li2022grounded,xu2022groupvit,ghiasi2022scaling} focused on combining object detection or semantic segmentation with various vision-language pretraining tasks, \textit{e.g.}, image-text contrastive learning~\cite{radford2021learning,yao2022filip} or image captioning~\cite{he2020image,nguyen2022grit}. 
However, these methods raise two questions, which may restrict the versatility of the above methods in universal tasks, \textit{e.g.}, universal image segmentation. 
\textbf{First}, are these methods~\cite{ren2015faster,zhang2020bridging,xu2022groupvit} feasible to handle universal tasks (\textit{e.g.}, semantic/instance/panoptic segmentation) simultaneously?
And \textbf{second}, could we leverage multiple training datasets with different kinds of annotations (\textit{e.g.}, both panoptic segmentation data~\cite{kirillov2019panoptic} and large-vocabulary detection data~\cite{shao2019objects365}) simultaneously to optimize such universal VL-based image segmentation framework? 

For the first question, the answer is negative, since the vanilla object detectors~\cite{ren2015faster,lin2017feature,zhang2020bridging} and semantic segmentation networks~\cite{long2015fully,xie2021segformer,xu2022groupvit} are not inherently feasible to handle universal segmentation tasks. 
Fortunately, some recent works regarding universal image segmentation with a single segmentation network~\cite{zhang2021knet,cheng2022masked,kmax_deeplab_2022} have been proposed, which inspires us to explore VL-based universal segmentation tasks based on these works.
And for the second question, intuitively, one can follow the preliminary exploration regarding omni-supervised learning methods~\cite{ren2020ufo,wang2022omni,chen2021points,yan2017weakly}, and introduce omni-supervised annotations into universal VL-based segmentation training. 
Benefiting from \textbf{1)} the high-quality things and stuff masks from panoptic segmentation datasets, and \textbf{2)} large-vocabulary visual concepts from object detection datasets and image-text pairs data, the VL-based models could obtain better open-world segmentation ability. 

To construct a universal open-world image segmentation method, in this paper, we start from a universal segmentation framework~\cite{cheng2022masked}, and propose \textit{Vision-Language Omni-Supervised Segmentation} (VLOSS), which is shown in Fig.~\ref{fig:overview}. 
Generally, VLOSS optimizes multiple VL-based recognition and segmentation tasks (\textit{i.e.}, panoptic segmentation~\cite{kirillov2019panoptic,li2021panopticfcn}, instance segmentation~\cite{he2017mask,cheng2022masked} and image-text pretraining task~\cite{radford2021learning}) in an end-to-end manner, thus obtaining the baseline structure of universal VL-based segmentation framework. 
To optimize all three tasks simultaneously within a single network, we first leverage the Mask2Former~\cite{cheng2022masked} as the image segmenter to extract object queries with corresponding segmentation masks for each region from things classes and stuff classes, and then introduce the CLIP text encoder to extract class embeddings for all class names and text embeddings for image captioning content. Then, the class embeddings are used to classify object queries and supervised by ground-truth labels, while the text embeddings are used to align with global image embeddings via image-text contrastive learning. 

Note that naive mixing of the training data with different tasks from different datasets leads to performance degeneration, since the lack of mask annotations from stuff classes in instance segmentation tasks may restrict the recognition ability of the segmentation network, thus affecting the overall performance on open-world segmentation benchmarks. 
To improve the training efficiency and release the power of omni-supervised training data, we propose several techniques to tackle this issue. 
First, we replace the deformable attention encoder in the original Mask2Former with FPN-style encoder, which aims to reduce the effect of overfitting from the original deformable encoder. This modification improves the open-world segmentation performance of VLOSS.
Second, we propose the switchable training technique (STT). During training, each training epoch is divided into warmup sub-epoch (for large-scale detection data) and cooldown sub-epoch (for panoptic segmentation data), such that VLOSS can preserve abundant visual concepts learned from detection data while obtaining recognition ability for stuff classes. 
And finally, we extend the training objectives, \ie, the classification loss $\mathcal{L}_{\text{cls}}$. For detection data, only the positive object queries matched by ground-truths are leveraged to calculate the classification loss, thus reducing the misclassification for stuff regions. 
After training, the optimized VLOSS can be adapted to different downstream open-world segmentation tasks without further adaptation. 

Extensive experiments have been conducted on two challenging open-world image segmentation benchmarks~\cite{zhou2017scene,gupta2019lvis}. Compared to the baselines, our method eliminates the negative effect of naive mixing the training data in different datasets, thus achieving better segmentation performance on multiple segmentation benchmarks. Moreover, even compared to previous state-of-the-art methods~\cite{ding2022open}, VLOSS also achieves comparable segmentation performance on open-world panoptic segmentation~\cite{zhou2017scene} and large-vocabulary instance segmentation~\cite{gupta2019lvis}. Notably, with 3$\times$ fewer parameters, VLOSS achieves comparable accuracy with MaskCLIP on ADE20K~\cite{zhou2017scene} and surpasses MaskCLIP by $\sim$2\% mask AP on LVIS v1.
%

\vspace{-0.5em}
\section{Related Work}
\vspace{-0.2em}
\subsection{Open-world Detection and Segmentation}\label{sec:related_owd_det_seg}
\vspace{-0.3em}
%
Open-world detection and segmentation aims to optimize a detection or segmentation network on a specific dataset to simultaneously localize and recognize both seen and unseen categories. 
Generally, current open-world detection and segmentation methods can be coarsely divided into two groups. 
The first is traditional class-agnostic detection and segmentation frameworks~\cite{qi2022open,xu2022dual,qi2022fine}, which aims to correctly localize all things and stuff regions via bounding boxes or masks, meanwhile recognizing the categories of regions belonging to known classes. 
And the second is vision-language open-world detection and segmentation methods~\cite{li2022grounded,ding2022open}, which usually leverage both \textbf{1)} an object detection~\cite{shao2019objects365} or segmentation dataset~\cite{lin2014microsoft,caesar2018coco} to construct fundamental localization ability and \textbf{2)} abundant image-text pair data to enhance the open-world recognition ability. 
Specifically, Li \textit{et al.} proposed grounded language-image pretraining (GLIP), which leveraged sufficient object detection data and massive image-text pairs to align a pretrained CLIP~\cite{radford2021learning} text encoder and an object detector, thus obtaining remarkable open-world detection or grounding ability on both seen and unseen categories. 
And Ding \textit{et al.} leveraged the off-the-shelf panoptic segmentation network~\cite{cheng2022masked} to extract class-agnostic segmentation masks, then optimized the newly proposed relative mask attention (RMA) module to extract the proposal embedding for each mask with corresponding images, finally classified the embedding via pretrained CLIP~\cite{radford2021learning}.
Our method belongs to the vision-language open-world segmentation method. 
%
%
%
In this paper, we will optimize a universal vision-language image segmentation network in an end-to-end manner and leverage massive omni-annotated data to enhance the universal image segmentation ability of our framework. 

\begin{figure*}
   \begin{center}
   \includegraphics[width=1.0\textwidth]{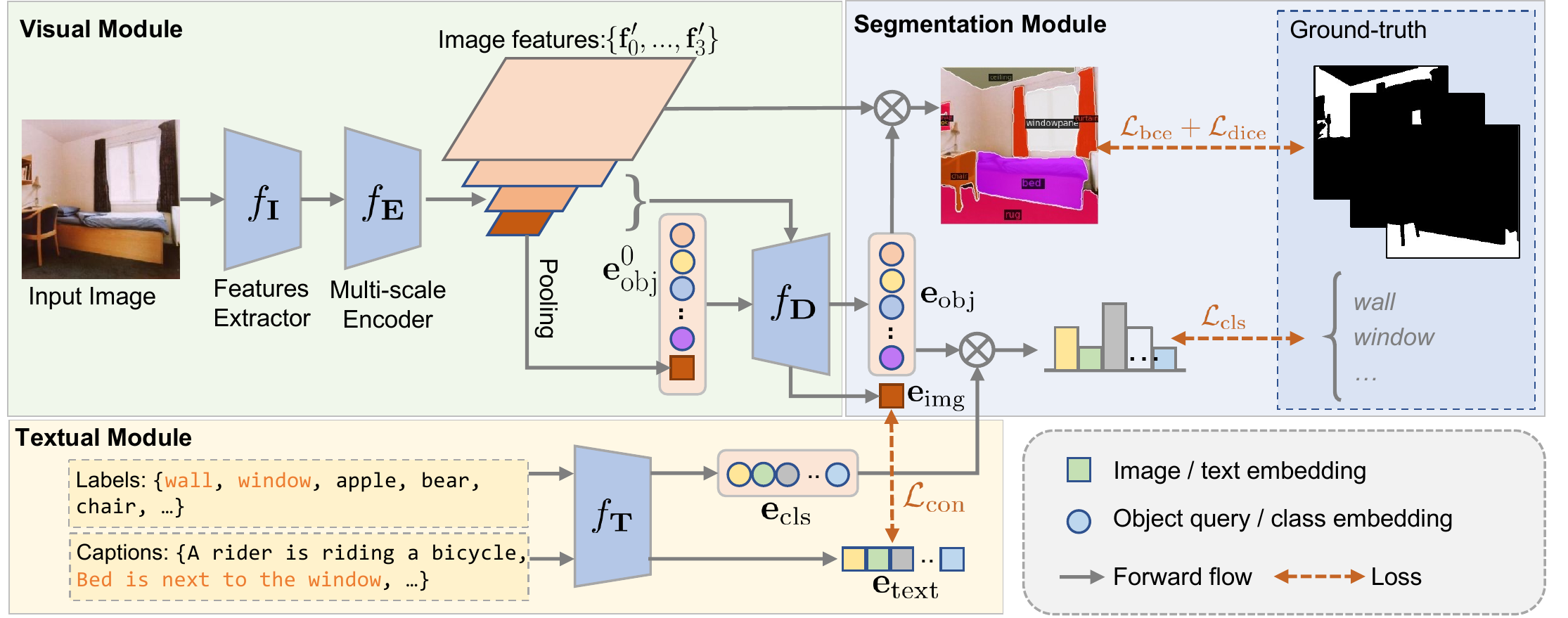}
   \end{center}
   \vspace{-1em}
   \caption{Overall architecture of our proposed VLOSS. With given input image, VLOSS uses feature extractor $f_{\mathbf{I}}$ and encoder $f_{\mathbf{E}}$ to extract multi-scale features. Then, the transformer-based decoder~\cite{cheng2022masked} leverages multi-scale features to obtain both object queries $\mathbf{e}_{\text{obj}}$ and global image embedding $\mathbf{e}_{\text{img}}$. With class embedding $\mathbf{e}_{\text{cls}}$ calculated by text encoder $f_{\mathbf{T}}$, VLOSS can obtain corresponding segmentation predictions. Meanwhile, the text embedding $\mathbf{e}_{\text{text}}$ from $f_{\mathbf{T}}$ are leveraged to align with $\mathbf{e}_{\text{img}}$ by image-text contrastive learning. Finally, VLOSS minimizes the loss function $\{\mathcal{L}_{\text{cls}},\mathcal{L}_{\text{mask}},\mathcal{L}_{\text{dice}}\}$ for panoptic/instance segmentation tasks and minimizes $,\mathcal{L}_{\text{con}}$ for image-text pretraining task.}
   \vspace{-1em}
   \label{fig:pipeline}
   \end{figure*}

\vspace{-0.5em}
\subsection{Omni-supervised Learning}\label{sec:related_omni}
\vspace{-0.5em}
Omni-supervised learning~\cite{ren2020ufo,wang2022omni,chen2021points,yan2017weakly} aims to optimize networks using both fully-supervised annotations and weakly-supervised annotations (\textit{e.g.}, using image labels or point annotations as supervision for object detection tasks) simultaneously during training, such that the learned models still obtain promising recognition ability against fully-supervised counterparts. 
%
%
Specifically, Ren \textit{et al.} strictly defined the definition of omni-supervised learning and proposed UFO$^2$ to construct a detector with various annotations. 
And Wang \textit{et al.} extended UFO$^2$ by a stronger object detector and more diverse annotations. 
Note that previous omni-supervised learning approaches mainly focus on a specific dataset (\textit{e.g.}, COCO~\cite{lin2014microsoft}) with multiple kinds of supervision. 
Yet seldom works manage to leverage multiple datasets with different kinds of supervision (\textit{e.g.}, COCO Panoptic segmentation dataset~\cite{lin2014microsoft,kirillov2019panoptic} with Objects365 detection dataset~\cite{shao2019objects365}). 
In this paper, we leverage omni-supervised annotations on different datasets simultaneously to optimize a universal VL-based segmentation framework. 

\section{Method}
\subsection{Overview of VLOSS}\label{sec:method_overview}
Fig.~\ref{fig:pipeline} shows the overall architecture of VLOSS. VLOSS coarsely includes two fundamental components, \textit{i.e.}, the Mask2Former-based~\cite{cheng2022masked} segmentation network as well as the text encoder~\cite{radford2021learning}. 
Specifically, with a given input image, we first feed this image into the image feature extractor $f_{\mathbf{I}}$ and obtain multi-scale image features $\left\{ \mathbf{f}_{0}, ..., \mathbf{f}_{3}\right\}$. Without loss of generality, we assume that $\mathbf{f}_{0}$ has the smallest spatial resolution and $\mathbf{f}_{3}$ has the largest resolution. Then, $\left\{ \mathbf{f}_{0}, ..., \mathbf{f}_{3}\right\}$ are fed into a multi-scale encoder $f_{\mathbf{E}}$ and obtain the refined multi-scale features as follows:
\begin{equation}
   \left\{ \mathbf{f}_{0}^{'}, ..., \mathbf{f}_{3}^{'}\right\} = f_\mathbf{E}(\left\{ \mathbf{f}_{0}, ..., \mathbf{f}_{3}\right\}).
\end{equation}
Next, given pre-defined initial object queries $\mathbf{e}^{0}_{\text{obj}}\in\mathbb{R}^{N\times D}$, where $N$ means the number of object queries and $D$ means the channel dimension of features and queries, we first obtain the global embedding by $\mathbf{e}^{0}_{\text{img}} = \phi(\mathbf{f}_{0}^{'})$, where $\phi$ means the average pooling operation. Then we obtain the refined object queries and global image embedding by multi-scale decoder $f_{\mathbf{D}}$~\cite{cheng2022masked} as follows:
\begin{equation}
   \left [ \mathbf{e}_{\text{obj}};\mathbf{e}_{\text{img}} \right ] = f_{\mathbf{D}}(\left [ \mathbf{e}^{0}_{\text{obj}};\mathbf{e}^{0}_{\text{img}} \right ], \left\{ \mathbf{f}_{0}^{'}, ..., \mathbf{f}_{2}^{'}\right\}).
\end{equation}
The obtained $\mathbf{e}_{\text{obj}}$ and $\mathbf{e}_{\text{img}}$ can be used to calculate and minimize the classification loss $\mathcal{L}_{\text{cls}}$ and contrastive loss $\mathcal{L}_{\text{con}}$, respectively. And finally, to achieve image segmentation tasks, one can obtain masks of object queries $\mathbf{M}$ by inner product between $\mathbf{e}_{\text{obj}}$ and $\mathbf{f}_{3}^{'}$.
Meanwhile, for text encoder $f_{\mathbf{T}}$, given class names $\left\{ \texttt{class\_name}\right\}$ from training sets, $f_{\mathbf{T}}$ first tokenizes each class name into word embeddings, then calculates the final class embeddings $\mathbf{e}_{\text{cls}}$ for classification.
A similar operation can be used on image captioning sentences and obtain $\mathbf{e}_{\text{text}}$ to calculate the contrastive loss.
Different from~\cite{li2022grounded}, we do not introduce the fusion module~\cite{dai2021dynamic,devlin2019bert} between image feature and text feature (\textit{e.g.}, $\mathbf{e}_{\text{cls}}$ or $\mathbf{e}_{\text{text}}$). The merit of this design is that: during open-world evaluation, the extracted class embedding can be reused, thus reducing the computation cost during evaluation and increasing flexibility. 
In the following, we first illustrate selected tasks, then start from Mask2Former~\cite{cheng2022masked} to build up the whole universal VL-based segmentation framework step by step. 
\subsection{Training Tasks in VLOSS}\label{sec:method_tasks}
To optimize the model in an end-to-end manner, without loss of generality, we introduce three kinds of typical tasks during training, \textit{i.e.}, panoptic segmentation, instance segmentation, and image-text contrastive learning. 
The reason why we chose these tasks is shown as follows.

\textbf{Panoptic Segmentation. }
Since we aim to construct a universal segmentation framework, \textit{i.e.}, it should segment and recognize each thing object and stuff region simultaneously. 
To introduce sufficient stuff class concepts and bring the fundamental segmentation ability into VLOSS, we choose panoptic segmentation as the main training task.

\textbf{Instance Segmentation. }
%
To introduce abundant visual concepts of things classes, we use Objects365~\cite{shao2019objects365} detection dataset, which owns hundreds of visual concepts to expand the label space. 
Note that we notice simply adopting omni-supervised learning with weakly-supervised segmentation loss~\cite{tian2021boxinst,hsu2019weakly} may lead to severe performance degeneration. 
Instead, for only training data, we use an off-the-shelf instance segmentation model~\cite{he2017mask} to extract pseudo masks for each bounding box and use the pseudo masks to ease the training difficulty.

\textbf{Image-text Pretraining Task. }
Recent works~\cite{li2022grounded,zhang2022glipv2} demonstrate that introducing image-text pairs data into open-world detection training can further improve the open-world recognition ability of models, since the highly structured and semantic-rich text data can enhance the representation ability of text encoder. 
During training, we also introduce image-text pretraining via contrastive learning as an auxiliary task. 

\subsection{Improving VLOSS from Architecture}\label{sec:method_encoder}
In the following, we start from the main training task, \textit{i.e.}, panoptic segmentation, to build up the baseline method. 
However, we notice that the baseline method performs limited on open-world segmentation benchmarks. We first investigate the network architecture and explore the bottleneck to tackle this issue. 
Based on previous works~\cite{radford2021learning,yang2022unified,zhu2021deformable}, we argue that a deformable transformer in the image encoder limits the open-world segmentation ability.
A feasible explanation is as follows: the deformable attention only encourages each pixel to fuse features from sparse and fixed number of pixels with high similarity. This design benefits locating and recognizing objects from seen categories while may not generalize well on unseen categories. 
To tackle this issue, we replace the deformable transformer encoder with an FPN-style encoder. Specifically, for the feature map with the smallest spatial resolution, we use vanilla transformer encoder~\cite{carion2020end} to extract features. 
And for features with a larger resolution, we follow FPN~\cite{lin2017feature} and use convolution layers to extract and fuse multi-scale features. 
Finally, multi-scale features are sent into VL-based decoder for segmentation and contrastive learning.

\subsection{Switchable Training Technique}\label{sec:method_stt}
\begin{figure}[t!]
\begin{center}
\includegraphics[width=0.46\textwidth]{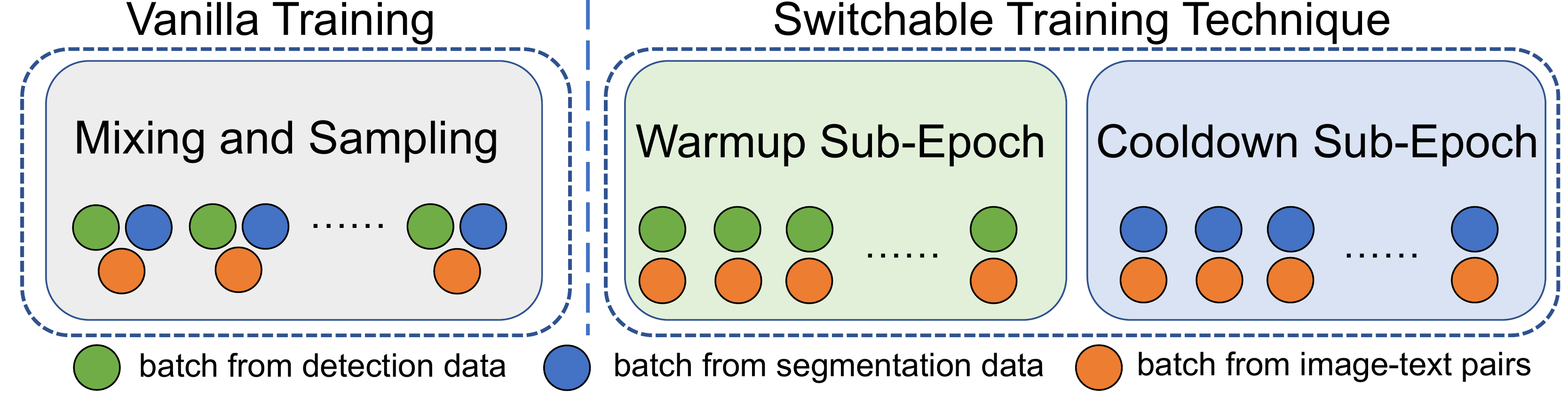}
\end{center}
\vspace{-1.5em}
\caption{Illustration of \textbf{(left)} an epoch with simply mixing all training data, and \textbf{(right)} an epoch of STT. The \textcolor{green}{green}, \textcolor{blue}{blue}, and \textcolor{orange}{orange} circles mean batches from detection datasets, panoptic segmentation datasets, and image-text pairs data, respectively.
}
\vspace{-1.5em}
\label{fig:stt}
\end{figure}

After improving the network architecture, one can explore the end-to-end training strategy of VLOSS. A straightforward idea is combining samples from all three tasks (as shown in Fig.~\ref{fig:stt} (left)), and it should obtain better segmentation performance than the baseline. 
However, when naive combines object detection data and panoptic segmentation data into a batch for training, we notice that the final performance could be better. Meanwhile, we also find that the image-text pretraining task can be compatible with any other task because this task only needs to align the global embedding between images and captions, which is easier than the other two dense prediction tasks.
A feasible explanation is that: 
\textbf{1)} for the same object may be categorized into different classes in different datasets (\textit{e.g.}, ``basketball'' in the detection dataset and ``sports ball'' in the segmentation dataset), which increases the difficulty of training; 
And \textbf{2)} object detection dataset lacks mask annotations of stuff region, such that for queries regarding stuff regions, these queries are assigned as ``no-object'' class, thus conflicting with the annotations in panoptic segmentation datasets. 
To tackle this issue, we propose a switchable training technique, which is shown in Fig.~\ref{fig:stt}. Instead of simply mixing all training data and random sampling, we split each training epoch into two sub-epochs, \textit{i.e.}, warmup sub-epoch, and cooldown sub-epoch. 
In warmup sub-epoch, both object detection datasets and image-text pairs data are leveraged to optimize the model. 
Benefiting from large-vocabulary detection data and semantic-rich captioning data, the model can be well initialized and obtain strong recognition ability for things classes. And during the cooldown epoch, both panoptic segmentation datasets and image-text pairs data are leveraged to fine-tune the model, thus further enhancing the recognition ability for both things classes and stuff classes. 
\subsection{Loss Functions}\label{sec:method_loss}
Finally, we introduce the training objectives to optimize the VLOSS training framework. Generally, during training of VLOSS, three kinds of losses (\textit{i.e.}, classification loss, segmentation loss, and image-text contrastive loss are leveraged to optimize the model simultaneously. Here we illustrate the details of these training objectives. 

\noindent\textbf{Segmentation Loss. }To ensure that each query can generate high-quality segmentation masks from dense image features, we also follow~\cite{cheng2022masked} and minimize the segmentation loss. 
Generally, we adopt the segmentation loss on panoptic segmentation and instance segmentation tasks. 
For each mask generated by the corresponding query, following~\cite{carion2020end,cheng2022masked}, we first leverage Hungarian matching~\cite{carion2020end} to match positive queries with corresponding masks, then we minimize binary cross-entropy loss~\cite{he2017mask} $\mathcal{L}_{\text{bce}}$ and Dice loss~\cite{milletari2016v} $\mathcal{L}_{\text{dice}}$ on masks from positive queries to optimize the model.

\noindent\textbf{Classification Loss. }To ensure that our method can obtain the fundamental region recognition ability, we leverage the classification loss for each query on panoptic and instance segmentation tasks. 
Specifically, for object queries $\mathbf{e}_{\text{obj}}$ as well as the class embeddings $\mathbf{e}_{\text{cls}}$ extracted by text encoder, where $\mathbf{e}_{\text{obj}}\in\mathbb{R}^{N\times D}$ includes $N$ object queries and $\mathbf{e}_{\text{cls}}\in\mathbb{R}^{(C+1)\times D}$ includes $C$ things or stuff categories as well as one pre-defined ``no-object'' category, we first calculate the per-class logits of queries $\mathbf{c}$ by $\mathbf{c} = \mathbf{e}_{\text{obj}}\mathbf{e}_{\text{cls}}^{T}$. Then we adopt Hungarian matching~\cite{carion2020end} to match each query with ground-truths (namely positive queries) or pre-defined ``no-object'' category (namely negative queries) and assign classification label $\mathbf{y}$ for $N$ queries. Finally, we use cross-entropy loss as the classification loss $\mathcal{L}_{\text{cls}}$ as follows:
\begin{equation}
   \mathcal{L}_{\text{cls}} = \frac{1}{N}\sum_{i=1}^{N}H(\mathbf{c}_{i}, \mathbf{y}_{i}),
\end{equation}
where $H$ means the cross-entropy function. 

Note that during instance segmentation tasks on object detection datasets, only objects from things classes are annotated. Thus for queries that localize the stuff regions, corresponding ground-truth labels are assigned as a pre-defined ``no-object'' class, which leads to misclassification, finally affecting the convergence speed of VLOSS and recognition ability for stuff classes. To reduce the negative effect, different from ~\cite{carion2020end,cheng2022masked}, for instance segmentation tasks, we only calculate $\mathcal{L}_{\text{cls}}$ on positive queries (namely positive classification loss), thus reduce the negative effect from lack of stuff classes annotations.

\noindent\textbf{Contrastive Loss. }To optimize the image-text pretraining task via contrastive learning, we adopt contrastive loss~\cite{radford2021learning,yao2022filip} between global embeddings of images and captions as an auxiliary training objective. Here we focus on a mini-batch with a batch size of $B$ to discuss this loss. 
Specifically, for global embedding of the $i$-th image $\mathbf{e}_{\text{img}}^{i}$ and global embedding of the $j$-th caption $\mathbf{e}_{\text{text}}^{j}$ ($1\leq i,j \leq B$), we first calculate the cosine similarity $\mathbf{s}_{i,j}$ between $\mathbf{e}_{\text{img}}^{i}$ and $\mathbf{e}_{\text{text}}^{j}$ as follows:
\begin{equation}
   \mathbf{s}_{i,j} = \langle \mathbf{e}_{\text{img}}^{i},\mathbf{e}_{\text{text}}^{j}\rangle / \tau,
\end{equation}
where $\langle\cdot,\cdot\rangle$ means cosine similarity operator and $\tau$ is the temperature term~\cite{radford2021learning}. Then, we follow~\cite{radford2021learning} to calculate the image-text contrastive loss $\mathcal{L}_{\text{con}}$ as Eq.~\ref{eq:clip_loss}: 
\begin{equation}\label{eq:clip_loss}
   \mathcal{L}_{\text{con}} = -\frac{1}{2B}\sum^{B}_{i=1}(\log\frac{\text{exp}(\mathbf{s}_{i,i})}{\Sigma^{B}_{j=1}\text{exp}({\mathbf{s}_{i,j}})} + \log\frac{\text{exp}(\mathbf{s}_{i,i})}{\Sigma^{B}_{j=1}\text{exp}({\mathbf{s}_{j,i}})}).
\end{equation}
We will discuss the selection of $\mathcal{L}_{\text{con}}$ in the ablation study. 
Finally, the training objective $\mathcal{L}$ is shown as follows:
\begin{equation}
   \mathcal{L} = \lambda_{\text{cls}}\mathcal{L}_{\text{cls}} + \lambda_{\text{bce}}\mathcal{L}_{\text{bce}} + \lambda_{\text{dice}}\mathcal{L}_{\text{dice}} + \lambda_{\text{con}}\mathcal{L}_{\text{con}}.
\end{equation}

\section{Experiments}
In the following, we summarize the training datasets we use for VLOSS, state the implementation details of VLOSS, then demonstrate the quantitative and qualitative results. Finally, we conduct the ablation study for VLOSS.
%

\begin{table*}[t]
   \caption{Comparison with state-of-the-art vision-language universal segmentation methods on ADE20K panoptic segmentation benchmark. Using fewer parameters, our VLOSS with Swin-Tiny backbone achieves comparable PQ$^{\text{all}}$ result than the state-of-the-art MaskCLIP, meanwhile achieving better PQ$^{\text{th}}$ than MaskCLIP. Top-2 results are bolded for better comparison.}
   \vspace{-1em}
   \label{table:ade20kpano}
   \begin{center}
      \setlength{\tabcolsep}{7.5pt} 
      \renewcommand{\arraystretch}{3.5}
              { \fontsize{8.3}{3.5}\selectfont{
   \begin{tabular}{l|c|c|c|c|ccc}
   \toprule
   {\bf Method} & {\bf Visual Enc} &{\bf Text Enc}&{\bf Params}& {\bf Proposal} &{\bf PQ$^{\text{all}}$} & {\bf PQ$^{\text{th}}$} & {\bf PQ$^{\text{st}}$} \\ 
   \midrule
   CLIP baseline & CLIP ViT-L & CLIP Text & 472.39M & \checkmark & 8.207 & 8.473 & 7.675  \\
   Mask2Former+CLIP & Swin-T & CLIP Text & \textbf{110.97M} & - & 11.003 & 8.932 & 15.144 \\
   MaskCLIP (Mask R-CNN)~\cite{ding2022open} & CLIP ViT-L & CLIP Text & 472.39M & \checkmark & 12.860 & 11.242 & 16.095 \\
   MaskCLIP~\cite{ding2022open} & CLIP ViT-L & CLIP Text & 472.03M & \checkmark & \textbf{15.121} & \textbf{13.536} & \textbf{18.290}  \\
   \midrule
   VLOSS & Swin-T &CLIP Text & \textbf{156.69M} & - & \textbf{15.371} &\textbf{13.959}& \textbf{18.195} \\
   \bottomrule
   \end{tabular}}}
   \vspace{-1em}
   \end{center}
   \end{table*}

\begin{table*}[t]
   \caption{Comparison with state-of-the-art on LVIS v1 \emph{val} and \emph{minival} set. Compared to the state-of-the-art MaskCLIP, VLOSS with Swin-T surpasses MaskCLIP by 1.8\% in terms of mask AP on LVIS v1 \emph{val} set, which demonstrate the effectiveness of VLOSS. Top-2 results are bolded for better comparison.}
   \vspace{-1em}
   \label{table:lvisv1ins}
   \begin{center}
      \setlength{\tabcolsep}{7pt} 
      \renewcommand{\arraystretch}{3.5}
               { \fontsize{8.3}{3.5}\selectfont{
   \begin{tabular}{l|c|c|c|c|ccc|ccc}
   \toprule
   {\bf Method} & {\bf Visual Enc} &{\bf Text Enc}&{\bf Params}& {\bf Proposal} &{\bf AP} & {\bf AP$^{\text{50}}$} & {\bf AP$^{\text{75}}$} & {\bf AP$_{\text{r}}$} & {\bf AP$_{\text{c}}$} & {\bf AP$_{\text{f}}$}\\ 
   \midrule
   \multicolumn{11}{l}{\bf Evaluated on \emph{val} set} \\
   \midrule
   CLIP baseline~\cite{radford2021learning} & CLIP ViT-L & CLIP Text & 472.39M & \checkmark & 5.0 & 7.2 & 5.2 & - & - & - \\
   Mask2Former+CLIP~\cite{cheng2022masked} & Swin-T & CLIP Text & \textbf{110.97M} & - & 5.3 & 7.8 & 5.6 & 2.1 & 3.1 & 9.1 \\
   MaskCLIP (Mask R-CNN)~\cite{ding2022open} & CLIP ViT-L & CLIP Text & 472.39M & \checkmark & 6.4 &\textbf{12.8} & 5.8 &- &- & - \\
   MaskCLIP~\cite{ding2022open} & CLIP ViT-L & CLIP Text & 472.03M & \checkmark & \textbf{8.4} &12.2 & \textbf{8.8} &- &- & - \\
   \midrule
   VLOSS & Swin-T &CLIP Text & \textbf{156.69M} & - & \textbf{10.2} & \textbf{15.8} & \textbf{10.6} & \textbf{4.4} & \textbf{8.1} & \textbf{15.0} \\
   \midrule
   \multicolumn{11}{l}{\bf Evaluated on \emph{minival} set} \\
   \midrule
   Mask2Former+CLIP~\cite{cheng2022masked} & Swin-T & CLIP Text & \textbf{110.97M} & - & 8.3 & 12.7 & 8.6 & 3.9 & 5.4 & 11.6 \\
   \midrule
   VLOSS & Swin-T &CLIP Text & \textbf{156.69M} & - & \textbf{13.8} & \textbf{20.7} & \textbf{14.7} & \textbf{8.1} & \textbf{12.3} & \textbf{16.2} \\
   \bottomrule
   \end{tabular}}}
   \vspace{-1em}
   \end{center}
   \end{table*}
\subsection{Training Datasets}\label{sec:exp_datasets}
\textbf{COCO Panoptic}~\cite{kirillov2019panoptic,lin2014microsoft} contains $\sim$118K and 5K images for training and validation, respectively. In this dataset, 80 things classes and 53 stuff classes are defined to evaluate the performance of panoptic segmentation. During experiments, we train the model on the \textit{train} set and select the best checkpoint on the \textit{val} set for further open-world evaluation. 

\textbf{Objects365}~\cite{shao2019objects365} includes $\sim$2M images and 365 individual object categories. During experiments, we randomly sample $\sim$0.66M training images from original training data since using sampled data is more efficient during training and can demonstrate the effectiveness of our method. 

\textbf{Image-Text Pairs Data. }During experiments, we use COCO caption~\cite{chen2015microsoft} to provide high-quality image-text pairs data, which includes $\sim$0.5M high-quality human-annotated image captioning sentences. We also introduce CC3M~\cite{sharma2018conceptual} into training, which will be discussed in the appendix.




\subsection{Implementation Details}\label{sec:exp_implement}
We use ImageNet~\cite{deng2009imagenet} pretrained Swin Transformer~\cite{liu2021swin} (\textit{i.e.}, Swin-Tiny) and the CLIP pretrained text encoder weights to initialize parameters of VLOSS. 
We do not use the CLIP-pretrained image encoder weights, and our experimental results show that even if without well-aligned visual and language encoders pair, our method can still achieve promising results. 
Following Mask2Former~\cite{cheng2022masked}, we set the number of object queries to 100 and use 9 pixel decoders to compute segmentation masks. 
And for efficient training, we reduce the maximum length of input text tokens from 77~\cite{radford2021learning} to 48 to reduce the computation cost during training, and introduce mixed precision during training to further control the computation cost meanwhile speedup training procedure. 
We use AdamW optimizer~\cite{loshchilov2018decoupled} to optimize the network. For panoptic segmentation and instance segmentation tasks, we set the batch size to 32 and use the input resolution of 768$\times$768. And for image-text pretraining task, we set the batch size to 512 and use the input resolution of 224$\times$224. 
Following previous works~\cite{cheng2022masked,radford2021learning}, we set $\lambda_{\text{cls}},\lambda_{\text{bce}},\lambda_{\text{dice}}$ to 2.0, 5.0, and 5.0. And we set $\lambda_{\text{con}}$ as 1.0 for simplicity. 
During training, we set the initial learning rate to 2.0$\times$10$^{\text{-4}}$, except for the text encoder, which is initialized with 2.0$\times$10$^{\text{-5}}$. Without otherwise specified, all models are trained with 12 epochs and the learning rate is decayed with a factor of 0.1 at the 8-th and the 11-th epoch.
All experiments are implemented by PyTorch toolkit~\cite{pytorch} and MMDetection code-base~\cite{chen2019mmdetection}, and all experiments are conducted on NVIDIA V100 GPUs.

\subsection{Comparison with State-of-The-Arts}\label{sec:exp_sota}
We select two challenging open-world segmentation tasks, \ie, open-world panoptic segmentation and open-world large-vocabulary instance segmentation, to evaluate the performance of VLOSS. 
The reason why we choose these two tasks is that, compared to traditional open-world semantic segmentation tasks~\cite{ding2022open,xu2022semseg,liu2022open}, these two tasks further require \textbf{1)} recognizing both things and stuff classes, and \textbf{2)} handling large-vocabulary instance-level recognition.
We compare our VLOSS with \textbf{1)} CLIP baseline, \textbf{2)} MaskCLIP~\cite{ding2022open}, and \textbf{2)} the state-of-the-art MaskCLIP~\cite{ding2022open}, which leverage the pretrained CLIP (ViT-L/14) as well as off-the-shelf COCO~\cite{lin2014microsoft} pretrained Mask R-CNN~\cite{he2017mask} to tackle open-world segmentation tasks.

\noindent\textbf{Open-world Panoptic Segmentation. }
Universal frameworks should simultaneously segment all objects and background regions from different (seen and unseen) things and stuff classes. Hence we first evaluate VLOSS on ADE20K panoptic segmentation benchmark~\cite{zhou2017scene} with state-of-the-art methods~\cite{ding2022open,cheng2022masked,radford2021learning}. 
Table~\ref{table:ade20kpano} demonstrates the performance of all the above methods. Our VLOSS significantly surpasses vanilla Mask2former with CLIP text encoder and CLIP with external proposal masks on a large margin. And surprisingly, even without well-aligned VL-pretrained weights~\cite{radford2021learning} and massive model parameters~\cite{dosovitskiy2021an}, our VLOSS with Swin-Tiny~\cite{liu2021swin} backbone achieves 15.371 in terms of PQ metric on all categories, which is comparable with ViT-L/14-based~\cite{dosovitskiy2021an} MaskCLIP with 3$\times$ more parameters. Moreover, our VLOSS also surpasses MaskCLIP~\cite{ding2022open} by $\sim$0.5\% in terms of PQ metric on things categories. The above results demonstrate the effectiveness of our proposed VLOSS on panoptic segmentation. 

\noindent\textbf{Open-world Instance Segmentation. }
To evaluate the performance of VLOSS on open-world large-vocabulary instance segmentation benchmarks, we compare our methods with previous state-of-the-art universal segmentation frameworks on LVIS v1 benchmark~\cite{gupta2019lvis} with 1203 classes. Following~\cite{ding2022open}, we report the open-world segmentation results on \emph{val} set, and we also report the evaluation results on LVIS v1 \emph{minival} set.
Note that we do not make comparisons with previous works~\cite{zhang2022glipv2,li2022grounded,yao2022detclip} which leverage Mask R-CNN~\cite{he2017mask} or ATSS~\cite{zhang2020bridging} as visual backbones since Mask2Former inherently performs unsatisfying on LVIS benchmarks, which will be discussed in the appendix. 
The evaluation results are shown in Table~\ref{table:lvisv1ins}. Notably, our VLOSS with Swin-Tiny largely surpasses ViT-L-based MaskCLIP by 1.8\% and 3.6\% in terms of mask AP and AP50 metrics, respectively. Especially, VLOSS also surpasses MaskCLIP by 3.7\% and 4.2\% in terms of mask AP on rare and medium classes, respectively. We also evaluate VLOSS on LVIS v1 \emph{minival} set, and VLOSS achieves 13.8\% and 20.7\% in terms of mask AP and AP50. The above results demonstrate the effectiveness of VLOSS.

\begin{figure*}
\begin{center}
\includegraphics[width=0.99\textwidth]{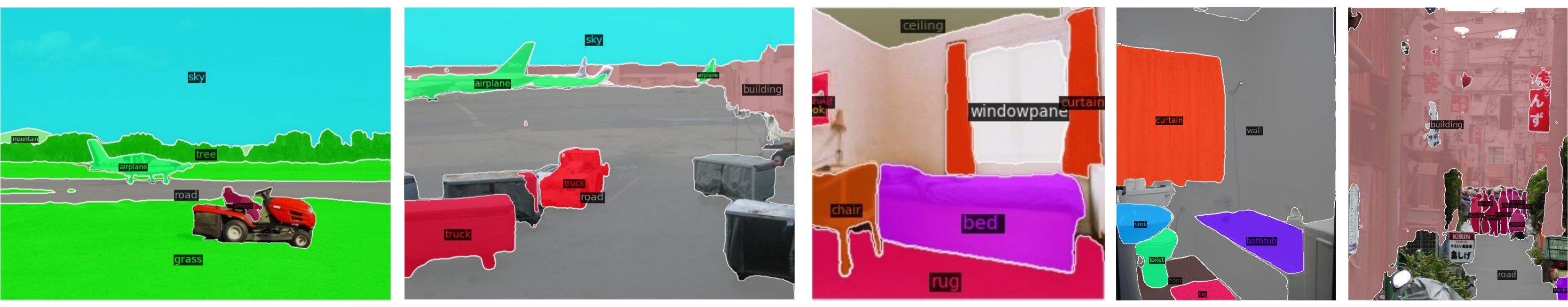}
\end{center}
\vspace{-1.5em}
\caption{Qualitative results of VLOSS on ADE20K panoptic segmentation benchmark. Both seen classes (\eg, ``sky'' and ``truck'') and unseen classes (\eg, ``windowpane'' and ``bathtub'') are segmented and recognized.}
\label{fig:ade20k_vis}
\end{figure*}
\begin{figure*}
\begin{center}
\includegraphics[width=0.99\textwidth]{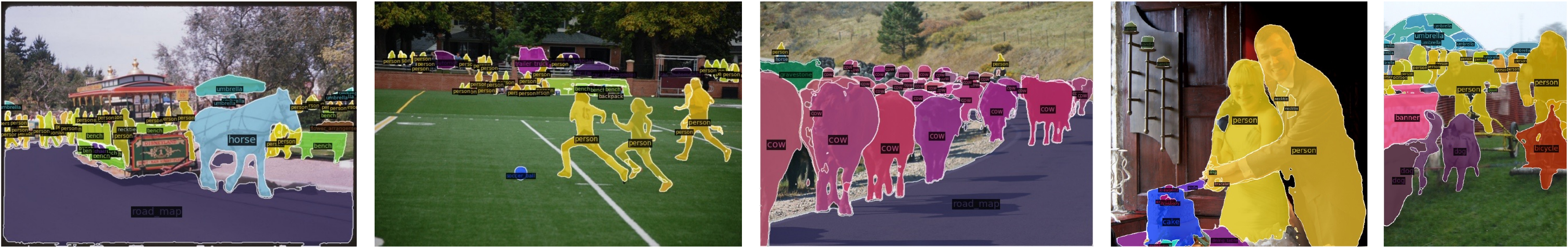}
\end{center}
\vspace{-1.5em}
\caption{Qualitative results of VLOSS on LVIS v1 instance segmentation benchmark. Both seen classes (\eg, ``person'' and ``cow'') and unseen classes (\eg, ``strawberry'' and ``necktie'') are segmented and recognized.}
\vspace{-1em}
\label{fig:lvisv1_vis}
\end{figure*}
\subsection{Qualitative Analysis}\label{sec:exp_vis}
To conduct a qualitative analysis of VLOSS, we demonstrate the visualization results of predictions from LVIS v1 dataset~\cite{gupta2019lvis} and ADE20K dataset~\cite{zhou2017scene}. Qualitative results of VLOSS are shown in Fig.~\ref{fig:ade20k_vis} and Fig.~\ref{fig:lvisv1_vis}, respectively. VLOSS can generate masks for both things and stuff categories for panoptic segmentation. Notably, for large-vocabulary instance segmentation, VLOSS still generates masks for objects from both seen and unseen (\eg, strawberry/necktie in Fig.~\ref{fig:lvisv1_vis}) categories. These results illustrate the effectiveness of VLOSS.

\begin{table}[t]
   \caption{Factor-to-factor ablation study of VLOSS, where ``O365'' means adding Objects365 to conduct omni-supervised training, ``pos $\mathcal{L}_{\text{cls}}$'' means calculating classification loss for only positive queries, ``STT'' means switchable training strategy, and $\mathcal{L}_{\text{con}}$ means adding image-text pairs data and contrastive loss into VLOSS.}
   \vspace{-1em}
   \label{table:ablation_overall}
   \begin{center}
      \setlength{\tabcolsep}{4.5pt} 
      \renewcommand{\arraystretch}{3.5}
              { \fontsize{8.3}{3}\selectfont{
   \begin{tabular}{ccccc|cc}
   \toprule
   {\bf Mask2Former} & {\bf O365} & {\bf pos $\mathcal{L}_{\text{cls}}$} & {\bf STT}  &{\bf $\mathcal{L}_{\text{con}}$}&{\bf PQ$^{\text{all}}$} & {\bf PQ$^{\text{th}}$}\\ 
   \midrule
   \checkmark & & & & & 11.003 & 8.932 \\
   \checkmark & \checkmark & & & & 9.816 & 10.641 \\
   \checkmark & \checkmark & \checkmark & & & 10.225 & 10.823 \\
   \checkmark & \checkmark & \checkmark & \checkmark & & 13.761 & 13.435 \\
   \midrule
   \checkmark & \checkmark & \checkmark  & \checkmark & \checkmark & \textbf{15.371} & \textbf{13.959} \\
   \bottomrule
   \end{tabular}}}
   \vspace{-2em}
   \end{center}
   \end{table}
\subsection{Ablation Study}\label{sec:exp_ablation}
Here we show ablation studies of VLOSS. Without specially mentioned, all ablation studies are evaluated on ADE20K open-world panoptic segmentation benchmark. 

\noindent\textbf{Factor-to-factor ablation study. }First, we conduct a factor-to-factor ablation study to analyze each component of VLOSS. 
The evaluation results are shown in Table~\ref{table:ablation_overall}. Specifically, after combining with CLIP text encoder, vanilla Mask2Former achieves 11.0\% PQ$^{\text{all}}$ metric. 
After simply introducing Objects365 into training, the PQ$^{\text{all}}$ decreases to 9.8\%, while PQ$^{\text{th}}$ increases to 10.6. 
These results show that introducing large-scale detection data as omni-supervised learning data indeed benefits open-world segmentation for things classes. However, without ``pos $\mathcal{L}_{\text{cls}}$'' and STT, the stuff regions are still misclassified, thus reducing PQ$^{\text{all}}$ metric.
%
After restricting the classification loss onto positive queries for instance segmentation task (\ie, ``pos $\mathcal{L}_{\text{cls}}$''), the PQ$^{\text{all}}$ increases to 10.2\%. 
Then, when further using STT to handle omni-supervised training, VLOSS achieves 13.8\% and 13.4\% in terms of PQ$^{\text{all}}$ and PQ$^{\text{th}}$, respectively. Finally, the image-text pairs data and contrastive loss making the final VLOSS achieve 15.4\% PQ$^{\text{all}}$.

\begin{table}[t]
   \caption{Ablation study of switchable training technique (STT). Results show that using STT performs better.}
   \vspace{-1.5em}
   \label{table:ablation_stt}
   \begin{center}
      \setlength{\tabcolsep}{7pt} 
      \renewcommand{\arraystretch}{3.5}
              { \fontsize{8.3}{3}\selectfont{
   \begin{tabular}{l|ccc|c}
   \toprule
   {\bf Method} & {\bf Mix} & {\bf Pretrain} &{\bf STT}&{\bf PQ$^{\text{all}}$}\\ 
   \midrule
w/ Mixing & \checkmark & & & 10.225 \\
    w/ Pretrain & & \checkmark & & 13.914 \\
   \midrule
    w/ STT & & & \checkmark & \textbf{15.371} \\
   \bottomrule
   \end{tabular}}}
   \vspace{-2em}
   \end{center}
   \end{table}

\noindent\textbf{Switchable Training Technique. }
Then we explore the effect of the switchable training technique. Here we select two baseline methods for comparison: 1) pretraining the network on a large-scale object detection dataset and then fine-tuning on panoptic segmentation dataset~\cite{chen2021pix2seq}, and 2) mixing all the training data during training. Experimental results are shown in Table~\ref{table:ablation_stt}. Compared to mixing all the training data, VLOSS using STT significantly outperforms VLOSS using mixing by 4.2\% in terms of PQ$^{\text{all}}$. Moreover, to illustrate that the proposed STT rather than Objects365~\cite{shao2019objects365} pretraining improves the segmentation ability, we also compare our method with fine-tuning VLOSS from an Objects365-pretrained backbone (namely VLOSS w/ pretrain). As shown in Table~\ref{table:ablation_stt}, VLOSS with STT still surpasses VLOSS using pretrain by 1.5\% in terms of PQ$^{\text{all}}$. An explanation is that: naive fine-tuning on panoptic segmentation may make VLOSS overfit to corresponding panoptic segmentation data, thus losing the large-vocabulary recognition ability from detection data. At the same time, our STT can tackle this issue since the design of STT ensures that large-vocabulary visual concepts can be revised in each epoch. 
The above results show the effectiveness of STT.

\begin{table}[t]
   \caption{Ablation study of differnet multi-scale encoder design in VLOSS. Using FPN-style encoder performs better.}
   \vspace{-1.5em}
   \label{table:ablation_encoder}
   \begin{center}
      \setlength{\tabcolsep}{7pt} 
      \renewcommand{\arraystretch}{3.5}
              { \fontsize{8.3}{3}\selectfont{
   \begin{tabular}{l|ccc}
   \toprule
   {\bf Method}&{\bf PQ$^{\text{all}}$}&{\bf PQ$^{\text{th}}$}&{\bf PQ$^{\text{st}}$}\\ 
   \midrule
    w/ Deform Encoder & 11.865 & 8.851 & 17.895 \\
    w/ FPN Encoder & 11.169 & 8.218 & 17.071 \\
    w/ FPN-style Encoder & \textbf{12.680} & \textbf{9.809} & \textbf{18.421} \\
   \bottomrule
   \end{tabular}}}
   \vspace{-1em}
   \end{center}
   \end{table}

\noindent\textbf{Effect of Encoder Design. }
As mentioned in Sec.~\ref{sec:method_encoder}, the deformable transformer encoder in~\cite{cheng2022masked} may restrict the open-world segmentation ability. To verify the effectiveness of the FPN-style encoder, we conduct an ablation study regarding the encoder in VLOSS. All experiments are trained on COCO panoptic and COCO captioning datasets. The results are shown in Table~\ref{table:ablation_encoder}. VLOSS using our FPN-style encoder outperforms VLOSS with deformable transformer encoder by 0.8\% and 1.0\% in terms of PQ$^{\text{all}}$ and PQ$^{\text{th}}$, respectively. Besides, we also compare our FPN-style encoder using vanilla FPN~\cite{lin2014microsoft}. VLOSS using FPN-style encoder surpasses VLOSS using FPN by 1.5\% PQ$^{\text{all}}$.
An explanation is that: both fully-convolutional-based FPN and deformable transformer encoder lack the global and long-range feature interaction, while FPN-style stated in Sec.~\ref{sec:method_encoder} can leverage vanilla transformer encoder to achieve this purpose.

\begin{table}[t]
   \caption{Ablation study of $\mathcal{L}_{\text{con}}$. Experimental results show that using CLIP loss as $\mathcal{L}_{\text{con}}$ performs better. }
   \vspace{-1em}
   \label{table:ablation_l_con}
   \begin{center}
      \setlength{\tabcolsep}{6.3pt} 
      \renewcommand{\arraystretch}{3.5}
              { \fontsize{8.3}{3}\selectfont{
   \begin{tabular}{l|cc|ccc}
   \toprule
   {\bf Method} & {\bf CLIP} &{\bf FILIP}&{\bf PQ$^{\text{all}}$} & {\bf PQ$^{\text{th}}$} & {\bf PQ$^{\text{st}}$}\\ 
   \midrule
    w/ FILIP &  & \checkmark & 10.669 & 8.246 & 15.515 \\
    w/ CLIP & \checkmark & & \textbf{12.680} &\textbf{9.809}& \textbf{18.421} \\
   \bottomrule
   \end{tabular}}}
   \vspace{-2em}
   \end{center}
   \end{table}

\noindent\textbf{Type of contrastive Loss. }
Finally, we discuss the choice of contrastive loss $\mathcal{L}_{\text{con}}$ in VLOSS. Intuitively, using better image-text contrastive loss as auxiliary loss can lead to better open-world segmentation performance. Hence we choose CLIP~\cite{radford2021learning} and FILIP~\cite{yao2022filip} to evaluate the corresponding effect. Without loss of generality, we optimize both VLOSS models on COCO panoptic and COCO captioning datasets. Experimental results are shown in Table~\ref{table:ablation_l_con}. Surprisingly, though FILIP pretraining usually obtain better VL-pretrained backbones than CLIP, VLOSS using CLIP still outperforms the counterpart by 2\% and 1.6\% in terms of PQ$^{\text{all}}$ and PQ$^{\text{st}}$ metrics, respectively. An explanation is that: FILIP aims to align the best-matched image patch with the corresponding text token. FILIP increases the training difficulty compared to vanilla CLIP loss, thus usually obtaining worse results than VLOSS using CLIP loss. Hence we choose CLIP loss as $\mathcal{L}_{\text{con}}$ in VLOSS.
\section{Conclusion}\label{sec:conclusion}
This paper demonstrates a universal open-world segmentation framework, namely Vision-Language Omni-Supervised Segmentation (VLOSS). 
   VLOSS starts from a Mask2Former universal segmentation framework with CLIP text encoder and leverages large-scale omni-supervised data (\ie, panoptic segmentation data, object detection data, and image-text pairs data) to enhance the open-world recognition ability, thus achieving promising open-world segmentation. To improve the training efficiency and fully release the power of omni-supervised data, we propose several advanced techniques, \ie, FPN-style encoder, switchable training technique, and positive classification loss.  
   %
   %
   Experimental results on different open-world panoptic and instance segmentation benchmarks demonstrate the effectiveness of VLOSS. 

\paragraph{Limitation.}
Our method still has two limitations. 
First, we only leverage bounding boxes and captions to optimize the model for weakly annotated datasets. Nevertheless, some regions that do not appear in the annotations (\textit{e.g.}, pixels from stuff classes) are not fully utilized during training. How to leverage the corresponding region via a self-supervised manner is worth discussing.
And second, we do not introduce visual grounding datasets (\textit{e.g.}, Ref-COCO~\cite{yu2016modeling} or Visual Genome~\cite{krishna2017visual}) into training, which can be used to improve open-world recognition ability further. 
Recent works~\cite{li2022grounded,zhang2022glipv2} have shown that introducing phrase grounding annotations can obtain more vital open-vocabulary recognition ability, which motivates us to leverage sufficient grounding data to enhance the open-world segmentation ability of VLOSS. 
In the future, we will focus on these limitations to further explore universal vision-language segmentation frameworks. 
{\small
\bibliographystyle{ieee_fullname}
\bibliography{egbib}

\begin{thebibliography}{10}\itemsep=-1pt

\bibitem{caesar2018coco}
Holger Caesar, Jasper Uijlings, and Vittorio Ferrari.
\newblock Coco-stuff: Thing and stuff classes in context.
\newblock In {\em Proceedings of the IEEE conference on computer vision and
  pattern recognition}, pages 1209--1218, 2018.

\bibitem{carion2020end}
Nicolas Carion, Francisco Massa, Gabriel Synnaeve, Nicolas Usunier, Alexander
  Kirillov, and Sergey Zagoruyko.
\newblock End-to-end object detection with transformers.
\newblock In {\em Computer Vision--ECCV 2020: 16th European Conference,
  Glasgow, UK, August 23--28, 2020, Proceedings, Part I 16}, pages 213--229.
  Springer, 2020.

\bibitem{chen2019mmdetection}
Kai Chen, Jiaqi Wang, Jiangmiao Pang, Yuhang Cao, Yu Xiong, Xiaoxiao Li,
  Shuyang Sun, Wansen Feng, Ziwei Liu, Jiarui Xu, et~al.
\newblock Mmdetection: Open mmlab detection toolbox and benchmark.
\newblock {\em arXiv preprint arXiv:1906.07155}, 2019.

\bibitem{chen2021points}
Liangyu Chen, Tong Yang, Xiangyu Zhang, Wei Zhang, and Jian Sun.
\newblock Points as queries: Weakly semi-supervised object detection by points.
\newblock In {\em Proceedings of the IEEE/CVF Conference on Computer Vision and
  Pattern Recognition}, pages 8823--8832, 2021.

\bibitem{chen2021pix2seq}
Ting Chen, Saurabh Saxena, Lala Li, David~J Fleet, and Geoffrey Hinton.
\newblock Pix2seq: A language modeling framework for object detection.
\newblock {\em arXiv preprint arXiv:2109.10852}, 2021.

\bibitem{chen2015microsoft}
Xinlei Chen, Hao Fang, Tsung-Yi Lin, Ramakrishna Vedantam, Saurabh Gupta, Piotr
  Doll{\'a}r, and C~Lawrence Zitnick.
\newblock Microsoft coco captions: Data collection and evaluation server.
\newblock {\em arXiv preprint arXiv:1504.00325}, 2015.

\bibitem{cheng2022masked}
Bowen Cheng, Ishan Misra, Alexander~G Schwing, Alexander Kirillov, and Rohit
  Girdhar.
\newblock Masked-attention mask transformer for universal image segmentation.
\newblock In {\em Proceedings of the IEEE/CVF Conference on Computer Vision and
  Pattern Recognition}, pages 1290--1299, 2022.

\bibitem{dai2021dynamic}
Xiyang Dai, Yinpeng Chen, Bin Xiao, Dongdong Chen, Mengchen Liu, Lu Yuan, and
  Lei Zhang.
\newblock Dynamic head: Unifying object detection heads with attentions.
\newblock In {\em Proceedings of the IEEE/CVF conference on computer vision and
  pattern recognition}, pages 7373--7382, 2021.

\bibitem{deng2009imagenet}
Jia Deng, Wei Dong, Richard Socher, Li-Jia Li, Kai Li, and Li Fei-Fei.
\newblock Imagenet: A large-scale hierarchical image database.
\newblock In {\em 2009 IEEE conference on computer vision and pattern
  recognition}, pages 248--255. Ieee, 2009.

\bibitem{devlin2019bert}
Jacob Devlin, Ming-Wei Chang, Kenton Lee, and Kristina Toutanova.
\newblock Bert: Pre-training of deep bidirectional transformers for language
  understanding.
\newblock In {\em Proceedings of the 2019 Conference of the North American
  Chapter of the Association for Computational Linguistics: Human Language
  Technologies, Volume 1 (Long and Short Papers)}, pages 4171--4186, 2019.

\bibitem{ding2022open}
Zheng Ding, Jieke Wang, and Zhuowen Tu.
\newblock Open-vocabulary panoptic segmentation with maskclip.
\newblock {\em arXiv preprint arXiv:2208.08984}, 2022.

\bibitem{dosovitskiy2021an}
Alexey Dosovitskiy, Lucas Beyer, Alexander Kolesnikov, Dirk Weissenborn,
  Xiaohua Zhai, Thomas Unterthiner, Mostafa Dehghani, Matthias Minderer, Georg
  Heigold, Sylvain Gelly, Jakob Uszkoreit, and Neil Houlsby.
\newblock An image is worth 16x16 words: Transformers for image recognition at
  scale.
\newblock In {\em International Conference on Learning Representations}, 2021.

\bibitem{ghiasi2022scaling}
Golnaz Ghiasi, Xiuye Gu, Yin Cui, and Tsung-Yi Lin.
\newblock Scaling open-vocabulary image segmentation with image-level labels.
\newblock In {\em Computer Vision--ECCV 2022: 17th European Conference, Tel
  Aviv, Israel, October 23--27, 2022, Proceedings, Part XXXVI}, pages 540--557.
  Springer, 2022.

\bibitem{gu2022openvocabulary}
Xiuye Gu, Tsung-Yi Lin, Weicheng Kuo, and Yin Cui.
\newblock Open-vocabulary object detection via vision and language knowledge
  distillation.
\newblock In {\em International Conference on Learning Representations}, 2022.

\bibitem{gupta2019lvis}
Agrim Gupta, Piotr Dollar, and Ross Girshick.
\newblock Lvis: A dataset for large vocabulary instance segmentation.
\newblock In {\em Proceedings of the IEEE/CVF conference on computer vision and
  pattern recognition}, pages 5356--5364, 2019.

\bibitem{he2017mask}
Kaiming He, Georgia Gkioxari, Piotr Doll{\'a}r, and Ross Girshick.
\newblock Mask r-cnn.
\newblock In {\em Proceedings of the IEEE international conference on computer
  vision}, pages 2961--2969, 2017.

\bibitem{he2020image}
Sen He, Wentong Liao, Hamed~R Tavakoli, Michael Yang, Bodo Rosenhahn, and
  Nicolas Pugeault.
\newblock Image captioning through image transformer.
\newblock In {\em Proceedings of the Asian conference on computer vision},
  2020.

\bibitem{hsu2019weakly}
Cheng-Chun Hsu, Kuang-Jui Hsu, Chung-Chi Tsai, Yen-Yu Lin, and Yung-Yu Chuang.
\newblock Weakly supervised instance segmentation using the bounding box
  tightness prior.
\newblock {\em Advances in Neural Information Processing Systems}, 32, 2019.

\bibitem{jia2021scaling}
Chao Jia, Yinfei Yang, Ye Xia, Yi-Ting Chen, Zarana Parekh, Hieu Pham, Quoc Le,
  Yun-Hsuan Sung, Zhen Li, and Tom Duerig.
\newblock Scaling up visual and vision-language representation learning with
  noisy text supervision.
\newblock In {\em International Conference on Machine Learning}, pages
  4904--4916. PMLR, 2021.

\bibitem{kirillov2019panoptic}
Alexander Kirillov, Kaiming He, Ross Girshick, Carsten Rother, and Piotr
  Doll{\'a}r.
\newblock Panoptic segmentation.
\newblock In {\em Proceedings of the IEEE/CVF Conference on Computer Vision and
  Pattern Recognition}, pages 9404--9413, 2019.

\bibitem{krishna2017visual}
Ranjay Krishna, Yuke Zhu, Oliver Groth, Justin Johnson, Kenji Hata, Joshua
  Kravitz, Stephanie Chen, Yannis Kalantidis, Li-Jia Li, David~A Shamma, et~al.
\newblock Visual genome: Connecting language and vision using crowdsourced
  dense image annotations.
\newblock {\em International journal of computer vision}, 123:32--73, 2017.

\bibitem{li2022grounded}
Liunian~Harold Li, Pengchuan Zhang, Haotian Zhang, Jianwei Yang, Chunyuan Li,
  Yiwu Zhong, Lijuan Wang, Lu Yuan, Lei Zhang, Jenq-Neng Hwang, et~al.
\newblock Grounded language-image pre-training.
\newblock In {\em Proceedings of the IEEE/CVF Conference on Computer Vision and
  Pattern Recognition}, pages 10965--10975, 2022.

\bibitem{li2021panopticfcn}
Yanwei Li, Hengshuang Zhao, Xiaojuan Qi, Liwei Wang, Zeming Li, Jian Sun, and
  Jiaya Jia.
\newblock Fully convolutional networks for panoptic segmentation.
\newblock In {\em IEEE Conference on Computer Vision and Pattern Recognition
  (CVPR)}, 2021.

\bibitem{lin2017feature}
Tsung-Yi Lin, Piotr Doll{\'a}r, Ross Girshick, Kaiming He, Bharath Hariharan,
  and Serge Belongie.
\newblock Feature pyramid networks for object detection.
\newblock In {\em Proceedings of the IEEE conference on computer vision and
  pattern recognition}, pages 2117--2125, 2017.

\bibitem{lin2014microsoft}
Tsung-Yi Lin, Michael Maire, Serge Belongie, James Hays, Pietro Perona, Deva
  Ramanan, Piotr Doll{\'a}r, and C~Lawrence Zitnick.
\newblock Microsoft coco: Common objects in context.
\newblock In {\em Computer Vision--ECCV 2014: 13th European Conference, Zurich,
  Switzerland, September 6-12, 2014, Proceedings, Part V 13}, pages 740--755.
  Springer, 2014.

\bibitem{lin2022frozen}
Ziyi Lin, Shijie Geng, Renrui Zhang, Peng Gao, Gerard de Melo, Xiaogang Wang,
  Jifeng Dai, Yu Qiao, and Hongsheng Li.
\newblock Frozen clip models are efficient video learners.
\newblock In {\em Computer Vision--ECCV 2022: 17th European Conference, Tel
  Aviv, Israel, October 23--27, 2022, Proceedings, Part XXXV}, pages 388--404.
  Springer, 2022.

\bibitem{liu2022open}
Quande Liu, Youpeng Wen, Jianhua Han, Chunjing Xu, Hang Xu, and Xiaodan Liang.
\newblock Open-world semantic segmentation via contrasting and clustering
  vision-language embedding.
\newblock In {\em Computer Vision--ECCV 2022: 17th European Conference, Tel
  Aviv, Israel, October 23--27, 2022, Proceedings, Part XX}, pages 275--292.
  Springer, 2022.

\bibitem{liu2021swin}
Ze Liu, Yutong Lin, Yue Cao, Han Hu, Yixuan Wei, Zheng Zhang, Stephen Lin, and
  Baining Guo.
\newblock Swin transformer: Hierarchical vision transformer using shifted
  windows.
\newblock In {\em Proceedings of the IEEE/CVF international conference on
  computer vision}, pages 10012--10022, 2021.

\bibitem{long2015fully}
Jonathan Long, Evan Shelhamer, and Trevor Darrell.
\newblock Fully convolutional networks for semantic segmentation.
\newblock In {\em Proceedings of the IEEE conference on computer vision and
  pattern recognition}, pages 3431--3440, 2015.

\bibitem{loshchilov2018decoupled}
Ilya Loshchilov and Frank Hutter.
\newblock Decoupled weight decay regularization.
\newblock In {\em International Conference on Learning Representations}, 2019.

\bibitem{milletari2016v}
Fausto Milletari, Nassir Navab, and Seyed-Ahmad Ahmadi.
\newblock V-net: Fully convolutional neural networks for volumetric medical
  image segmentation.
\newblock In {\em 2016 fourth international conference on 3D vision (3DV)},
  pages 565--571. Ieee, 2016.

\bibitem{nguyen2022grit}
Van-Quang Nguyen, Masanori Suganuma, and Takayuki Okatani.
\newblock Grit: Faster and better image captioning transformer using dual
  visual features.
\newblock In {\em Computer Vision--ECCV 2022: 17th European Conference, Tel
  Aviv, Israel, October 23--27, 2022, Proceedings, Part XXXVI}, pages 167--184.
  Springer, 2022.

\bibitem{pytorch}
Adam Paszke, Sam Gross, Francisco Massa, Adam Lerer, James Bradbury, Gregory
  Chanan, Trevor Killeen, Zeming Lin, Natalia Gimelshein, Luca Antiga, Alban
  Desmaison, Andreas Kopf, Edward Yang, Zachary DeVito, Martin Raison, Alykhan
  Tejani, Sasank Chilamkurthy, Benoit Steiner, Lu Fang, Junjie Bai, and Soumith
  Chintala.
\newblock Pytorch: An imperative style, high-performance deep learning library.
\newblock In {\em Advances in Neural Information Processing Systems 32}, pages
  8024--8035. Curran Associates, Inc., 2019.

\bibitem{qi2022fine}
Lu Qi, Jason Kuen, Weidong Guo, Tiancheng Shen, Jiuxiang Gu, Wenbo Li, Jiaya
  Jia, Zhe Lin, and Ming-Hsuan Yang.
\newblock Fine-grained entity segmentation.
\newblock {\em arXiv preprint arXiv:2211.05776}, 2022.

\bibitem{qi2022open}
Lu Qi, Jason Kuen, Yi Wang, Jiuxiang Gu, Hengshuang Zhao, Philip Torr, Zhe Lin,
  and Jiaya Jia.
\newblock Open world entity segmentation.
\newblock {\em IEEE Transactions on Pattern Analysis and Machine Intelligence},
  2022.

\bibitem{radford2021learning}
Alec Radford, Jong~Wook Kim, Chris Hallacy, Aditya Ramesh, Gabriel Goh,
  Sandhini Agarwal, Girish Sastry, Amanda Askell, Pamela Mishkin, Jack Clark,
  et~al.
\newblock Learning transferable visual models from natural language
  supervision.
\newblock In {\em International conference on machine learning}, pages
  8748--8763. PMLR, 2021.

\bibitem{rao2022denseclip}
Yongming Rao, Wenliang Zhao, Guangyi Chen, Yansong Tang, Zheng Zhu, Guan Huang,
  Jie Zhou, and Jiwen Lu.
\newblock Denseclip: Language-guided dense prediction with context-aware
  prompting.
\newblock In {\em Proceedings of the IEEE/CVF Conference on Computer Vision and
  Pattern Recognition}, pages 18082--18091, 2022.

\bibitem{ren2015faster}
Shaoqing Ren, Kaiming He, Ross Girshick, and Jian Sun.
\newblock Faster r-cnn: Towards real-time object detection with region proposal
  networks.
\newblock {\em Advances in neural information processing systems}, 28, 2015.

\bibitem{ren2020ufo}
Zhongzheng Ren, Zhiding Yu, Xiaodong Yang, Ming-Yu Liu, Alexander~G Schwing,
  and Jan Kautz.
\newblock Ufo 2: A unified framework towards omni-supervised object detection.
\newblock In {\em Computer Vision--ECCV 2020: 16th European Conference,
  Glasgow, UK, August 23--28, 2020, Proceedings, Part XIX}, pages 288--313.
  Springer, 2020.

\bibitem{shao2019objects365}
Shuai Shao, Zeming Li, Tianyuan Zhang, Chao Peng, Gang Yu, Xiangyu Zhang, Jing
  Li, and Jian Sun.
\newblock Objects365: A large-scale, high-quality dataset for object detection.
\newblock In {\em Proceedings of the IEEE/CVF international conference on
  computer vision}, pages 8430--8439, 2019.

\bibitem{sharma2018conceptual}
Piyush Sharma, Nan Ding, Sebastian Goodman, and Radu Soricut.
\newblock Conceptual captions: A cleaned, hypernymed, image alt-text dataset
  for automatic image captioning.
\newblock In {\em Proceedings of the 56th Annual Meeting of the Association for
  Computational Linguistics (Volume 1: Long Papers)}, pages 2556--2565,
  Melbourne, Australia, July 2018. Association for Computational Linguistics.

\bibitem{tian2021boxinst}
Zhi Tian, Chunhua Shen, Xinlong Wang, and Hao Chen.
\newblock Boxinst: High-performance instance segmentation with box annotations.
\newblock In {\em Proceedings of the IEEE/CVF Conference on Computer Vision and
  Pattern Recognition}, pages 5443--5452, 2021.

\bibitem{wang2022omni}
Pei Wang, Zhaowei Cai, Hao Yang, Gurumurthy Swaminathan, Nuno Vasconcelos,
  Bernt Schiele, and Stefano Soatto.
\newblock Omni-detr: Omni-supervised object detection with transformers.
\newblock In {\em Proceedings of the IEEE/CVF conference on computer vision and
  pattern recognition}, pages 9367--9376, 2022.

\bibitem{wu2022end}
Mingrui Wu, Jiaxin Gu, Yunhang Shen, Mingbao Lin, Chao Chen, Xiaoshuai Sun, and
  Rongrong Ji.
\newblock End-to-end zero-shot hoi detection via vision and language knowledge
  distillation.
\newblock {\em arXiv preprint arXiv:2204.03541}, 2022.

\bibitem{xie2021segformer}
Enze Xie, Wenhai Wang, Zhiding Yu, Anima Anandkumar, Jose~M Alvarez, and Ping
  Luo.
\newblock Segformer: Simple and efficient design for semantic segmentation with
  transformers.
\newblock {\em Advances in Neural Information Processing Systems},
  34:12077--12090, 2021.

\bibitem{xu2022dual}
Hai-Ming Xu, Hao Chen, Lingqiao Liu, and Yufei Yin.
\newblock Dual decision improves open-set panoptic segmentation.
\newblock {\em The 33rd British Machine Vision Conference (BMVC) 2022}, 2022.

\bibitem{xu2022groupvit}
Jiarui Xu, Shalini De~Mello, Sifei Liu, Wonmin Byeon, Thomas Breuel, Jan Kautz,
  and Xiaolong Wang.
\newblock Groupvit: Semantic segmentation emerges from text supervision.
\newblock In {\em Proceedings of the IEEE/CVF Conference on Computer Vision and
  Pattern Recognition}, pages 18134--18144, 2022.

\bibitem{xu2022semseg}
Mengde Xu, Zheng Zhang, Fangyun Wei, Yutong Lin, Yue Cao, Han Hu, and Xiang
  Bai.
\newblock A simple baseline for open-vocabulary semantic segmentation
  with pre-trained vision-language model.
\newblock In Shai Avidan, Gabriel Brostow, Moustapha Ciss{\'e}, Giovanni~Maria
  Farinella, and Tal Hassner, editors, {\em Computer Vision -- ECCV 2022},
  pages 736--753, Cham, 2022. Springer Nature Switzerland.

\bibitem{yan2017weakly}
Ziang Yan, Jian Liang, Weishen Pan, Jin Li, and Changshui Zhang.
\newblock Weakly-and semi-supervised object detection with
  expectation-maximization algorithm.
\newblock {\em arXiv preprint arXiv:1702.08740}, 2017.

\bibitem{yang2022unified}
Jianwei Yang, Chunyuan Li, Pengchuan Zhang, Bin Xiao, Ce Liu, Lu Yuan, and
  Jianfeng Gao.
\newblock Unified contrastive learning in image-text-label space.
\newblock In {\em Proceedings of the IEEE/CVF Conference on Computer Vision and
  Pattern Recognition}, pages 19163--19173, 2022.

\bibitem{yao2022detclip}
Lewei Yao, Jianhua Han, Youpeng Wen, Xiaodan Liang, Dan Xu, Wei Zhang, Zhenguo
  Li, Chunjing Xu, and Hang Xu.
\newblock Det{CLIP}: Dictionary-enriched visual-concept paralleled pre-training
  for open-world detection.
\newblock In Alice~H. Oh, Alekh Agarwal, Danielle Belgrave, and Kyunghyun Cho,
  editors, {\em Advances in Neural Information Processing Systems}, 2022.

\bibitem{yao2022filip}
Lewei Yao, Runhui Huang, Lu Hou, Guansong Lu, Minzhe Niu, Hang Xu, Xiaodan
  Liang, Zhenguo Li, Xin Jiang, and Chunjing Xu.
\newblock {FILIP}: Fine-grained interactive language-image pre-training.
\newblock In {\em International Conference on Learning Representations}, 2022.

\bibitem{yu2016modeling}
Licheng Yu, Patrick Poirson, Shan Yang, Alexander~C Berg, and Tamara~L Berg.
\newblock Modeling context in referring expressions.
\newblock In {\em Computer Vision--ECCV 2016: 14th European Conference,
  Amsterdam, The Netherlands, October 11-14, 2016, Proceedings, Part II 14},
  pages 69--85. Springer, 2016.

\bibitem{kmax_deeplab_2022}
Qihang Yu, Huiyu Wang, Siyuan Qiao, Maxwell Collins, Yukun Zhu, Hartwig Adam,
  Alan Yuille, and Liang-Chieh Chen.
\newblock k-means mask transformer.
\newblock In {\em ECCV}, 2022.

\bibitem{zhang2022glipv2}
Haotian Zhang, Pengchuan Zhang, Xiaowei Hu, Yen-Chun Chen, Liunian~Harold Li,
  Xiyang Dai, Lijuan Wang, Lu Yuan, Jenq-Neng Hwang, and Jianfeng Gao.
\newblock Glipv2: Unifying localization and vision-language understanding.
\newblock In {\em Advances in Neural Information Processing Systems}, 2022.

\bibitem{zhang2022tip}
Renrui Zhang, Wei Zhang, Rongyao Fang, Peng Gao, Kunchang Li, Jifeng Dai, Yu
  Qiao, and Hongsheng Li.
\newblock Tip-adapter: Training-free adaption of clip for few-shot
  classification.
\newblock In {\em Computer Vision--ECCV 2022: 17th European Conference, Tel
  Aviv, Israel, October 23--27, 2022, Proceedings, Part XXXV}, pages 493--510.
  Springer, 2022.

\bibitem{zhang2020bridging}
Shifeng Zhang, Cheng Chi, Yongqiang Yao, Zhen Lei, and Stan~Z Li.
\newblock Bridging the gap between anchor-based and anchor-free detection via
  adaptive training sample selection.
\newblock In {\em Proceedings of the IEEE/CVF conference on computer vision and
  pattern recognition}, pages 9759--9768, 2020.

\bibitem{zhang2021knet}
Wenwei Zhang, Jiangmiao Pang, Kai Chen, and Chen~Change Loy.
\newblock {K-Net: Towards} unified image segmentation.
\newblock In {\em NeurIPS}, 2021.

\bibitem{zhao2022exploiting}
Shiyu Zhao, Zhixing Zhang, Samuel Schulter, Long Zhao, BG Vijay~Kumar,
  Anastasis Stathopoulos, Manmohan Chandraker, and Dimitris~N Metaxas.
\newblock Exploiting unlabeled data with vision and language models for object
  detection.
\newblock In {\em Computer Vision--ECCV 2022: 17th European Conference, Tel
  Aviv, Israel, October 23--27, 2022, Proceedings, Part IX}, pages 159--175.
  Springer, 2022.

\bibitem{zhou2017scene}
Bolei Zhou, Hang Zhao, Xavier Puig, Sanja Fidler, Adela Barriuso, and Antonio
  Torralba.
\newblock Scene parsing through ade20k dataset.
\newblock In {\em Proceedings of the IEEE conference on computer vision and
  pattern recognition}, pages 633--641, 2017.

\bibitem{zhou2022conditional}
Kaiyang Zhou, Jingkang Yang, Chen~Change Loy, and Ziwei Liu.
\newblock Conditional prompt learning for vision-language models.
\newblock In {\em Proceedings of the IEEE/CVF Conference on Computer Vision and
  Pattern Recognition}, pages 16816--16825, 2022.

\bibitem{zhou2022learning}
Kaiyang Zhou, Jingkang Yang, Chen~Change Loy, and Ziwei Liu.
\newblock Learning to prompt for vision-language models.
\newblock {\em International Journal of Computer Vision}, 130(9):2337--2348,
  2022.

\bibitem{zhu2021deformable}
Xizhou Zhu, Weijie Su, Lewei Lu, Bin Li, Xiaogang Wang, and Jifeng Dai.
\newblock Deformable {\{}detr{\}}: Deformable transformers for end-to-end
  object detection.
\newblock In {\em International Conference on Learning Representations}, 2021.

\end{thebibliography}
}

\end{document}